\documentclass[runningheads]{llncs}
\usepackage{graphicx}
\usepackage{csquotes}
\usepackage[hidelinks]{hyperref}
\usepackage{amsfonts, amssymb, amsmath}
\usepackage{isomath}
\usepackage[nameinlink]{cleveref}
\begin{document}
\title{Eigenpatches — Adversarial Patches from Principal Components}
\titlerunning{Eigenpatches - Adv. Patches from Principal Components}
%
\author{Jens Bayer\inst{1,2}\href{https://orcid.org/0000-0002-2806-6920}{\includegraphics[scale=0.04]{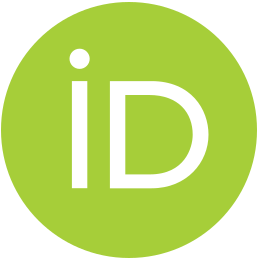}}\href{mailto:jens.bayer@iosb.fraunhofer.de}{\includegraphics[scale=0.04]{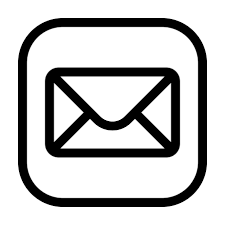}} \and
Stefan Becker\inst{1,2}\href{https://orcid.org/0000-0001-7367-2519}{\includegraphics[scale=0.04]{images/orcid-og-image.png}} \and
David Münch\inst{1,2}\href{https://orcid.org/0000-0002-8577-5256}{\includegraphics[scale=0.04]{images/orcid-og-image.png}} \and
Michael Arens\inst{1,2}\href{https://orcid.org/0000-0002-7857-0332}{\includegraphics[scale=0.04]{images/orcid-og-image.png}}
}
\authorrunning{J. Bayer et al.}
%
\institute{Fraunhofer Center for Machine Learning, \and
Fraunhofer IOSB,\\ Gutleuthausstr. 1, 76275 Ettlingen, Germany\\
\email{jens.bayer@iosb.fraunhofer.de}\\
}%
\maketitle              

\begin{abstract}
Adversarial patches are still a simple yet powerful white box attack that 
can be used to fool object detectors by suppressing possible detections.
The patches of these so-called evasion attacks are computational expensive 
to produce and require full access to the attacked detector. This paper 
addresses the problem of computational expensiveness by analyzing 375
generated patches, calculating the principal components of these and show,
that linear combinations of the resulting \enquote{eigenpatches} can be used 
to fool object detections successfully.

\end{abstract}
\section{Introduction}
Despite the good performance in image classification, object detection and semantic segmentation, computer 
vision systems are still prone to adversarial attacks. A prominent white box attack scenario is an evasion 
attack against object detectors using adversarial patches, whereas these patches are sometimes also 
referred to as a form of camouflage \cite{Duan2021} or invisibility cloak \cite{Wu2020,Zhu2022}. 
This is possible since the patches depict specifically calculated patterns, that alter the activations of the 
attacked object class for an object detector.

As \enquote{simple} and straightforward the generation process of such a patch is, as time-consuming and computation 
intensive it is as well. To speed up the patch calculation and gain a more profound understanding if there are 
common attributes that can be extracted to harden an object detector, this paper investigates:
\begin{enumerate}
    \item If patches that are combinations of principal components generated by a set of precalculated patches 
        can be used to create valid patches. 
    \item How many principal components are necessary, for recreating patches with a sufficient high impact on the
        detection score.
    \item The influence of set size used to generate the principal components, regarding the quality of the 
        resulting patches.
\end{enumerate}

\begin{figure}[!htb]
    \centering
    \includegraphics[width=0.45\textwidth]{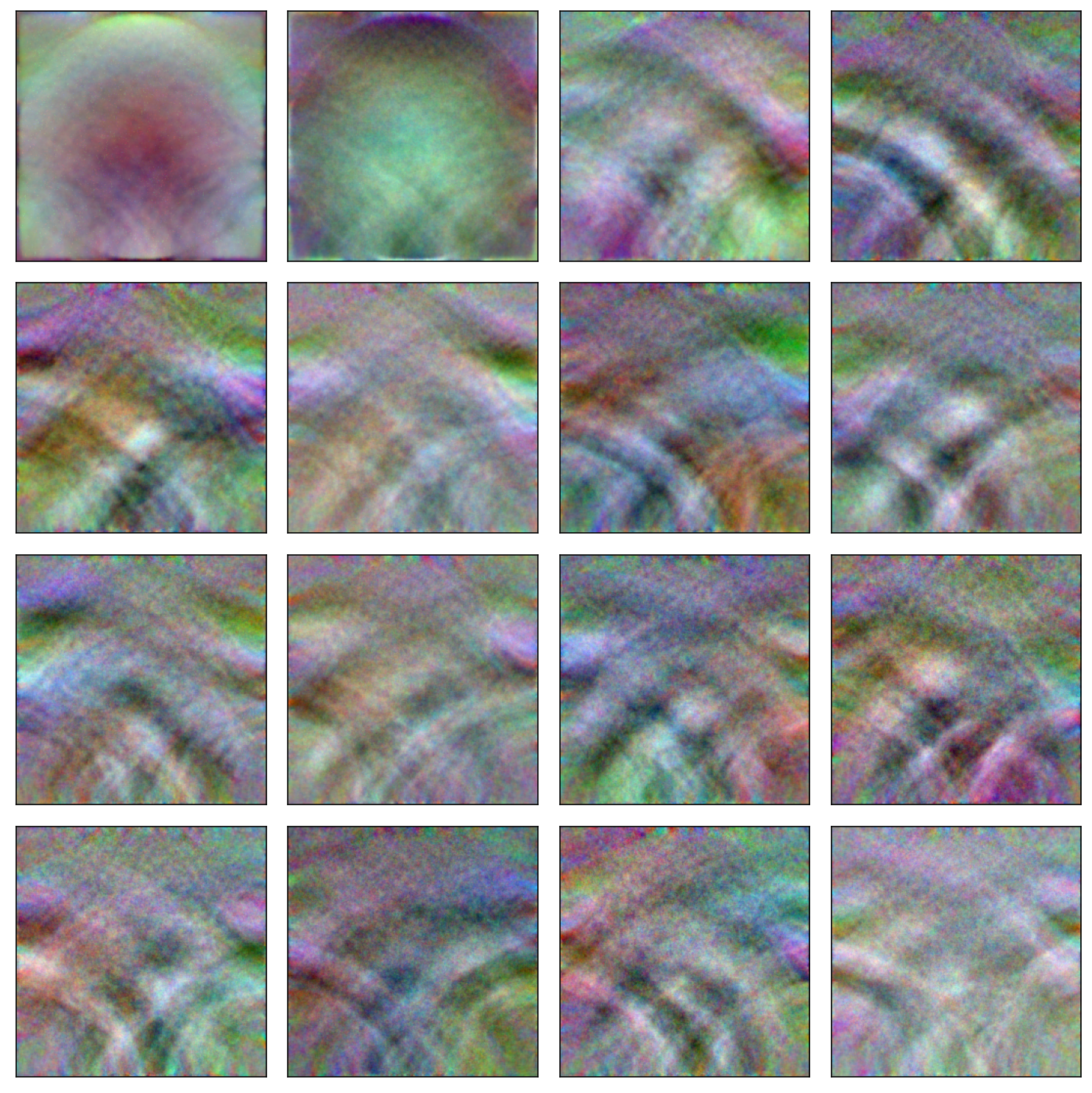}
    \includegraphics[width=0.45\textwidth]{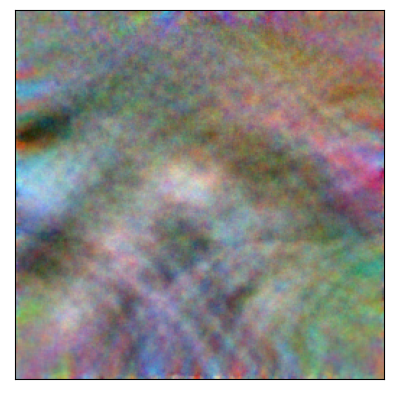}
    \caption{Left: the first sixteen principal components extracted from the trained patches. Right: the unweighted mean 
        image of these sixteen principal components.}
    \label{fig:meanpatch}
\end{figure}

Towards this end, the well known YOLOv7 object detector \cite{Wang2022} is attacked with various adversarial patches. These
patches are trained on a subset of the INRIA Person dataset \cite{Dalal2005} and are analyzed with a principal 
component analysis (PCA). The extracted principal components are then used to recreate the patches and even create new
patches by combining them linearly (e.g., \autoref{fig:meanpatch}).

To the best of our knowledge, this is the first approach in investigating and creating adversarial patches that attack 
object detectors by using principal components.

The organization of the paper is as follows: In \autoref{sec:related_work} the related work regarding the investigation and 
analysis of adversarial attacks is addressed. A brief introduction of what eigenpatches are and how they can be
crafted is given in \autoref{sec:eigenpatches}. \Cref{sec:experimental_setup} covers all necessary information to recreate
the experimental results. The used object detector, the dataset, and the generation process of the patches that are the base
for eigenpatch generation are further explained. The results of the experiments and the used metrics are presented in 
\autoref{sec:evaluation}. Finally, \autoref{sec:conclusion} summarizes the paper, gives a conclusion and a brief outlook.
%

\section{Related Work}
\label{sec:related_work}
In the following section, selected works regarding the analysis of adversarial attack patterns and sampling from lower 
dimensional embeddings are presented. For a broader overview, the interested reader is encouraged to read the survey 
papers for adversarial attacks \cite{Akhtar2018,Chakraborty2021,Pauling2022,Wang2023} and the YOLO object detector 
family \cite{Terven2023}.


The approach of using dimensionality reduction algorithms like principal component analysis to investigate the 
patterns generated by adversarial attacks for image classifiers is an active research area. Nonetheless, most 
investigations are performed solely on image classifiers with attacks, that induce high frequent noise on the whole image 
\cite{Tramer2017,Wang2019,Shi2021,Dohmatob2022,Shafahi2019a,Weng2023,Garcia2023}. Only a few works \cite{Tarchoun2023} investigate 
adversarial patches.

Tramer et al. explore the space of transferable adversarial examples by proposing methods for estimating the dimensionality of the space 
of adversarial inputs \cite{Tramer2017}. By investigating untargeted misclassification attacks, they show that perturbing a data point
in such a way, that it crosses a model's decision boundary, is likely to result in similar performance degradation when applied to other models.

Wang et al. present a fast black-box adversarial attack that finds key differences between different classes based on 
PCA that later can be used to drive a sample to a target class or the nearest other class \cite{Wang2019}.

Energy Attack \cite{Shi2021} is a transfer-based black-box $L_\infty$ adversarial attack that uses PCA to obtain the energy 
distribution of perturbations, generated by white-box attacks on a surrogate model. For the attack, patches are sampled 
from the energy distribution, tiled and applied on the target image.

Dohmatob et al. investigate, whether neural networks are vulnerable to black-box attacks inherently \cite{Dohmatob2022}. 
After analyzing low-dimensional adversarial perturbations, they hypothesize that adversarial perturbations exist with high 
probability in low dimensional subspaces, that are much smaller than the dimension of the image space.

Theoretical bounds on the vulnerability of classifiers against adversarial attacks are shown by Shafahi et al. using the unit sphere
and unit cube \cite{Shafahi2019a}. They claim that by using extremely large values for the class density functions, the bounds can be 
potentially escaped.

Weng et al. perform a Singular Value Decomposition to improve adversarial attacks on convolutional neural networks \cite{Weng2023}.
They combine the top-1 decomposed singular value-associated features for computing the output logits with the original logits, used to
optimize adversarial examples and thus boost the transferability of the attacks.

Recently, Tarchoun et al. examined adversarial patches from an information theory perspective and measured the
entropy of random crops of the patches \cite{Tarchoun2023}. The results indicate that the entropy of adversarial patches 
have a higher mean entropy than natural images. Based on these findings, they create a defense mechanism against
adversarial patches.

Further research on the relationship between adversarial vulnerability and the number of perturbed dimensions is 
performed by Godfrey et al. \cite{Godfrey2023}. Their results strengthen the hypothesis, that adversarial examples are a 
result of the locally linear behavior of neural networks with high dimensional input spaces.


\section{Eigenpatches}
\label{sec:eigenpatches}
This section covers the generation process of the eigenpatches. The term eigenpatches is derived from eigenfaces (the most 
prominent example of eigenpictures \cite{Sirovich1987}), which are calculated similarly. Eigenpictures is the name of the
eigenvectors, that can be derived from a set of training images. They can be used as a low-dimensional representation 
of the original training images and recreating these through a linear combination.

Given a set of adversarial patches 
\begin{equation}
    \mathcal{P} = \{ \tensorsym{P}_i | i = 1, \ldots, n \}, \quad \tensorsym{P}_i \in \left[0,1\right]^{C \times H \times W}
\end{equation}
where $H$ is the height in pixel, $W$ is the width in pixel and $C$ is the number of channels. A principal component analysis
is now performed on $\mathcal{P}$. With the top $k$ principal components $\tensorsym{E}_j, j \in \{1, \ldots, k\}$ and the weights
$\lambda_{i,j}$ the set 
\begin{equation}
    \mathcal{\hat{P}} = \{ \tensorsym{\hat{P}}_i | i = 1, \ldots, n\} \quad \tensorsym{\hat{P}}_i = \sum_{j=1}^k \lambda_{i,j} \tensorsym{E}_j
\end{equation}
can be generated, that consists of linear combinations of the principal components and is a recreation of $\mathcal{P}$.
\autoref{fig:meanpatch} shows the first sixteen principal components as well as the patch resulting from calculating the mean 
of the components.


\section{Experimental Setup}
\label{sec:experimental_setup}
In the following, all the necessary information to recreate the experimental results is presented. In addition,
the full source code is also provided\footnote{\url{https://github.com/JensBayer/PCAPatches}}. The attacked object 
detector is YOLOv7 \cite{Wang2020} with the official pretrained weights. The used Dataset is the INRIA Person 
dataset \cite{Dalal2005}.

\subsection{Object Detector}
Due to its prominence and good performance, YOLOv7 \cite{Wang2022} is selected as the object detector to be attacked. The 
smallest model of YOLOv7 with an input size of $640\times640$ pixels and the official pretrained weights are used. Despite
a small fix of an error, that causes problems during the training of the patches, the source code is not modified. The generated 
patches can therefore be used without specific finetuned weights, and the results can easily be verified.

\subsection{Dataset}
\label{sec:dataset}
\begin{figure}[!htb]
    \centering
    \includegraphics[width=0.5\textwidth]{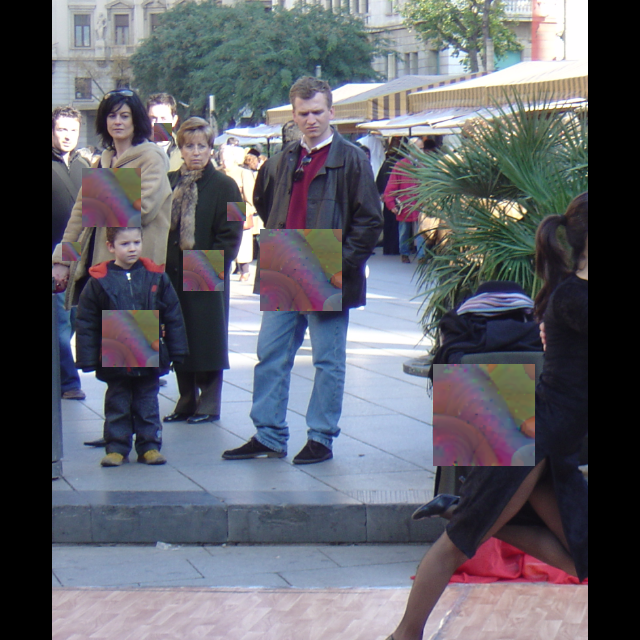}
    \caption{Patched input image as it would be used in the evaluation. The image is resized and padded to match the 
        required input size of $640\times640$ pixels. After this, the tested patch is embedded in the center of each 
        bounding box and resized according to a scaling factor.}
    \label{fig:patched_input_example}
\end{figure}

For the generation of the patches, the positive images of the INRIA Person dataset \cite{Dalal2005} are used. Instead
of using the provided ground truth bounding boxes of the dataset, the object detector first generates new 
ground truth data for each train image. After this, the newly created bounding boxes are manually reviewed.
False positives and bounding boxes of highly obscured target classes (persons) are removed. The resulting
ground truth data for both, the train and test split, are available in the source code repository. 
Since the input size of the used YOLOv7 model is $640\times640$ pixels, the images are resized and padded to 
match this size. 
A resized and patched example image as it would be used in the evaluation is given in \autoref{fig:patched_input_example}.

\subsection{Patch Generation}
\label{sec:patchgeneration}
The training procedure of the patches follows the procedure, described in \cite{Thys2019}:
During the generation process of a patch, it is placed inside the bounding boxes of selected target
classes in images of a training dataset. 
To improve the robustness of the patch, different randomized transformations, such as translation, rotation,
perspective and also color jitter are applied. 
The altered image is then propagated through the object detector and the objectness score of the target 
classes is minimized. In addition, a smoothness loss that punishes high frequent noise in the patches is 
also calculated and applied.

All patches are initialized with random values in the range $[0, 1]$ and optimized with the AdamW 
\cite{Loshchilov2019} optimizer with an initial learning rate of $0.01$. The random color jitter 
changes the brightness, contrast, and saturation of the patch randomly  by a values in the range 
of $[0.9, 1.1]$, $[0.95, 1.05]$, and $[0.97, 1.03]$ respectively. For the perspective, the distortion 
scale parameter is set to 0.5. The resize range of the patch, the learning rate scheduler and 
the number of training epochs are defined by the values in \autoref{tab:parameterization}. 

The random translation of the patch is performed in such a way, that the resized and transformed 
patch is placed inside the bounding box.

\begin{table}[!htb]
    \centering
    \begin{tabular}{|c|c|c|c|c|c|}
    \hline
         Run ID & n & Epochs & LR-Scheduler & Resize range & Rotation \\
         \hline
        191 & 175 & 100 & StepLR & [0.75, 1.0] & 30 \\
        885 & 100 & 100 & CosineAnnealingLR & [0.75, 1.0] & 30 \\
        905 & 50 & 100 & StepLR & [0.75, 1.0] & 45 \\
        371 & 25 & 125 & StepLR & [0.5, 0.75] & 30 \\
        243 & 25 & 125 & StepLR & [0.5, 0.75] & 45 \\
    \hline
    \end{tabular}
    \caption{Different parameterization for the patch generation. The resize range is the range of 
        the scaling factor, relatively to the bounding box.}
    \label{tab:parameterization}
\end{table}

\section{Evaluation}
\label{sec:evaluation}
This section covers the used metrics, the results, and a final limitations section 
that discusses, under which circumstances these results are generalizable
and valid they are. The $\uparrow$ and $\downarrow$ in the following tables 
indicate, whether a higher or lower value in the column is desired.

\subsection{Metrics}
To quantify the results, the well-known mean average precision object detector metrics 
mAP@.5 \cite{Everinghm2006} and mAP@.5:.95 \cite{Lin2014} are used. While mAP@.5 is 
the mean average precision for bounding boxes with an intersection over union (IoU) threshold 
of at least fifty percent, the mAP@.5:.95 is the average across ten IoU thresholds and
therefore more strict. Both metrics are calculated for the unpatched and the patched
images. The higher the difference between the unpatched and patched metric scores, the
stronger is the impact of the patches on the detector. 
For \autoref{tab:results_global}, the  differences ($\Delta$) between the unpatched and 
patched inputs images is also given. 

\subsection{Experimental Setup}
Similar to the training data, the test split and positive images of the INRIA Person dataset are used
for the evaluation. New ground truth bounding boxes are generated the same way as described in
\autoref{sec:dataset}. 
To generate reproducible results, the evaluation procedure is as follows: A patch is placed
in the center of each bounding box as given in the ground truth information and resized to 
a relative size of 0.75 times the longer side of the bounding box. Other transformations 
such as the ones mentioned in \autoref{sec:patchgeneration} are not applied. An example of 
a test image is given in \autoref{fig:patched_input_example}. After this, the
mAP@.5 and mAP@.5:.95 is calculated, according to the provided detector evaluation script.

\subsection{Results}
As shown in \autoref{tab:results_global}, the trained patches have a significant impact on
the object detector. While the mAP of the detector on the test data is about 0.73 (0.71), the
mAP drops to 0.46 (0.35) when attacked with the adversarial patches. To verify, that this is
not a result of covering the persons in the image with the patches, we also checked eleven 
single-colored gray valued patches. 
Each of these patches has one particular shade of gray, out of the uniformly distributed range from zero to one.
As the second row of \autoref{tab:results_global}
shows, the gray valued patches do not cover important parts and even improve the mAP@.5 slightly.
In comparison, the 375 crafted patches result in heavy mAP drops. 

\begin{figure}
    \centering
    \includegraphics[width=0.8\textwidth]{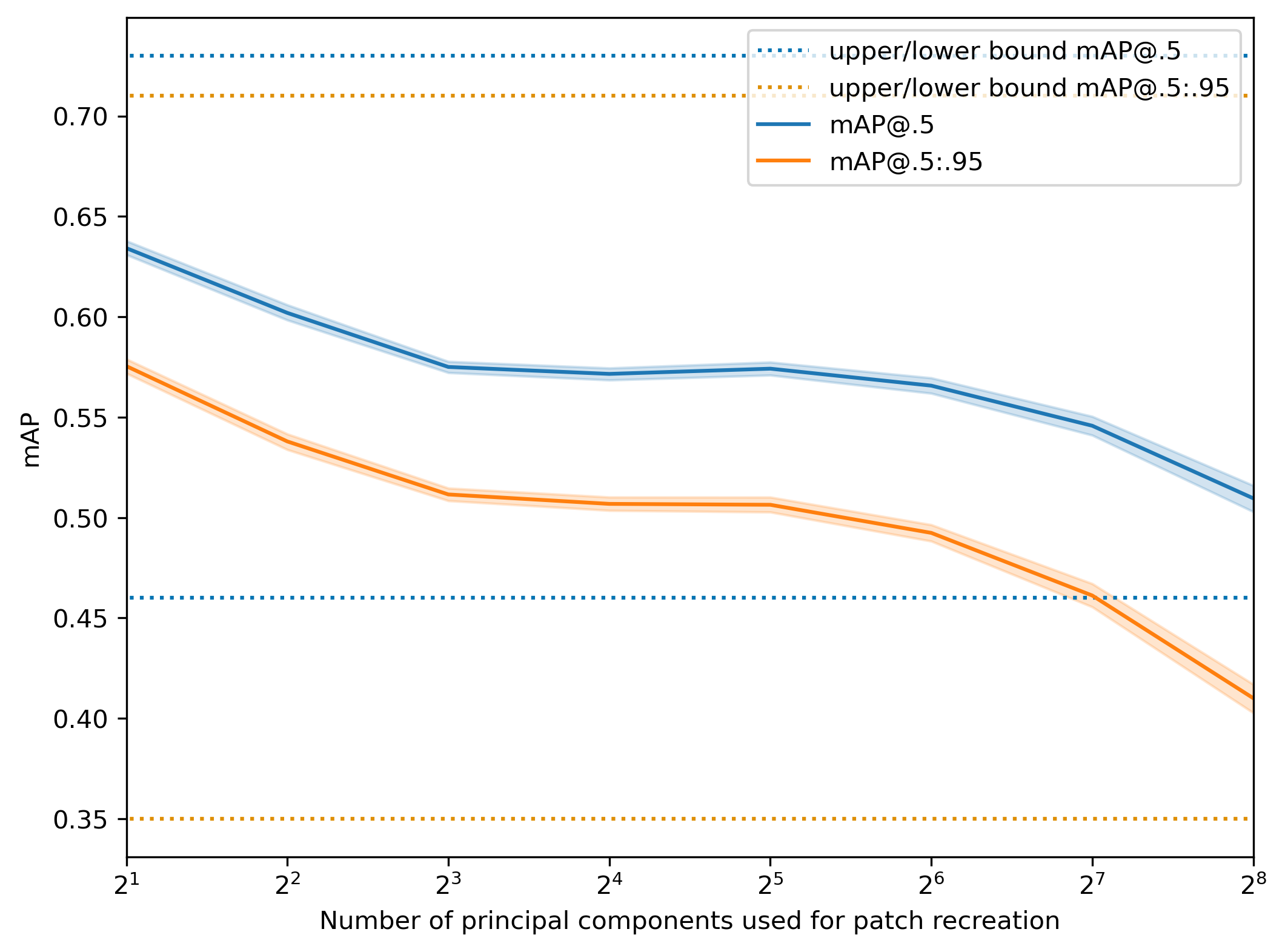}
    \caption{Drop of the mAP@.5 and mAP@.5:.95 with different numbers of principal components to the recovered patches.}
    \label{fig:mAPdrop}
\end{figure}

As expected, the PCA recovered patches are not as good as the trained ones and the more principal
components are used to recreate the patch, the higher the mAP drop is (see \autoref{fig:mAPdrop}). 
The rising but still small standard deviation suggests that some patches can be recreated 
better than others, which seems plausible, since the number of patches with the same 
parameterization as given in \autoref{tab:parameterization} is unevenly distributed. In addition, 
the mAP@.5 difference between the patches that use 128 principal components for recreation and the 
patches that only use eight principal components with about 0.03 is relatively small. Yet, the 
impact on the mAP@.5:.95 is almost twice as high (0.05).

    \begin{table}[!tb]
        \centering
        \begin{tabular}{|c|c||c|c||c|c|}
        \hline
             Applied Patches & Count & $mAP@.5 \downarrow$ & $mAP@.5:.95 \downarrow$ & $\Delta mAP@.5 \uparrow$ & $\Delta mAP@.5:.95 \uparrow$ \\
             \hline
             No & 1 & $0.73 $ & $0.71$ &  &  \\
             \hline
             Gray value & 11 & $0.74 \pm 0.03$ & $0.71 \pm 0.03$ & $-0.01\pm0.03$ & $0.00\pm0.02$ \\
             Trained & 375 & $\mathbf{0.46 \pm 0.03}$ & $\mathbf{0.35 \pm 0.03}$ & $\mathbf{0.27\pm0.03}$ & $\mathbf{0.36\pm0.03}$ \\
             \hline
             \hline
             PCA(2)   recovered & 375 & $0.63 \pm 0.03$ & $0.58 \pm 0.03$ & $0.10 \pm 0.03$ & $0.13 \pm 0.03$\\
             PCA(4)   recovered & 375 & $0.60 \pm 0.04$ & $0.54 \pm 0.04$ & $0.13 \pm 0.04$ & $0.17 \pm 0.04$\\
             PCA(8)   recovered & 375 & $0.58 \pm 0.03$ & $0.51 \pm 0.03$ & $0.15 \pm 0.03$ & $0.20 \pm 0.03$\\
             PCA(16)  recovered & 375 & $0.57 \pm 0.03$ & $0.51 \pm 0.03$ & $0.16 \pm 0.03$ & $0.20 \pm 0.03$\\
             PCA(32)  recovered & 375 & $0.57 \pm 0.03$ & $0.51 \pm 0.04$ & $0.16 \pm 0.03$ & $0.20 \pm 0.04$\\
             PCA(64)  recovered & 375 & $0.57 \pm 0.04$ & $0.49 \pm 0.04$ & $0.16 \pm 0.04$ & $0.22 \pm 0.04$\\
             PCA(128) recovered & 375 & $0.55 \pm 0.05$ & $0.46 \pm 0.06$ & $0.18 \pm 0.05$ & $0.25 \pm 0.06$\\
             PCA(256) recovered & 375 & $0.51 \pm 0.07$ & $0.41 \pm 0.07$ & $0.22 \pm 0.07$ & $0.30 \pm 0.07$\\

        \hline
        \end{tabular}
        \caption{Mean average precision for the unpatched and patched test data. 
            \enquote{Gray value} refers to gray values in the range of zero to 
            one with a linear step size. \enquote{Count} is the numbers of patches 
            on which the result mean and standard deviation values are calculated.
            The last two columns are the differences of the mAPs of the patched
            and unpatched images.}
        \label{tab:results_global}
    \end{table}

\begin{figure}
    \centering
    \includegraphics[width=0.8\textwidth]{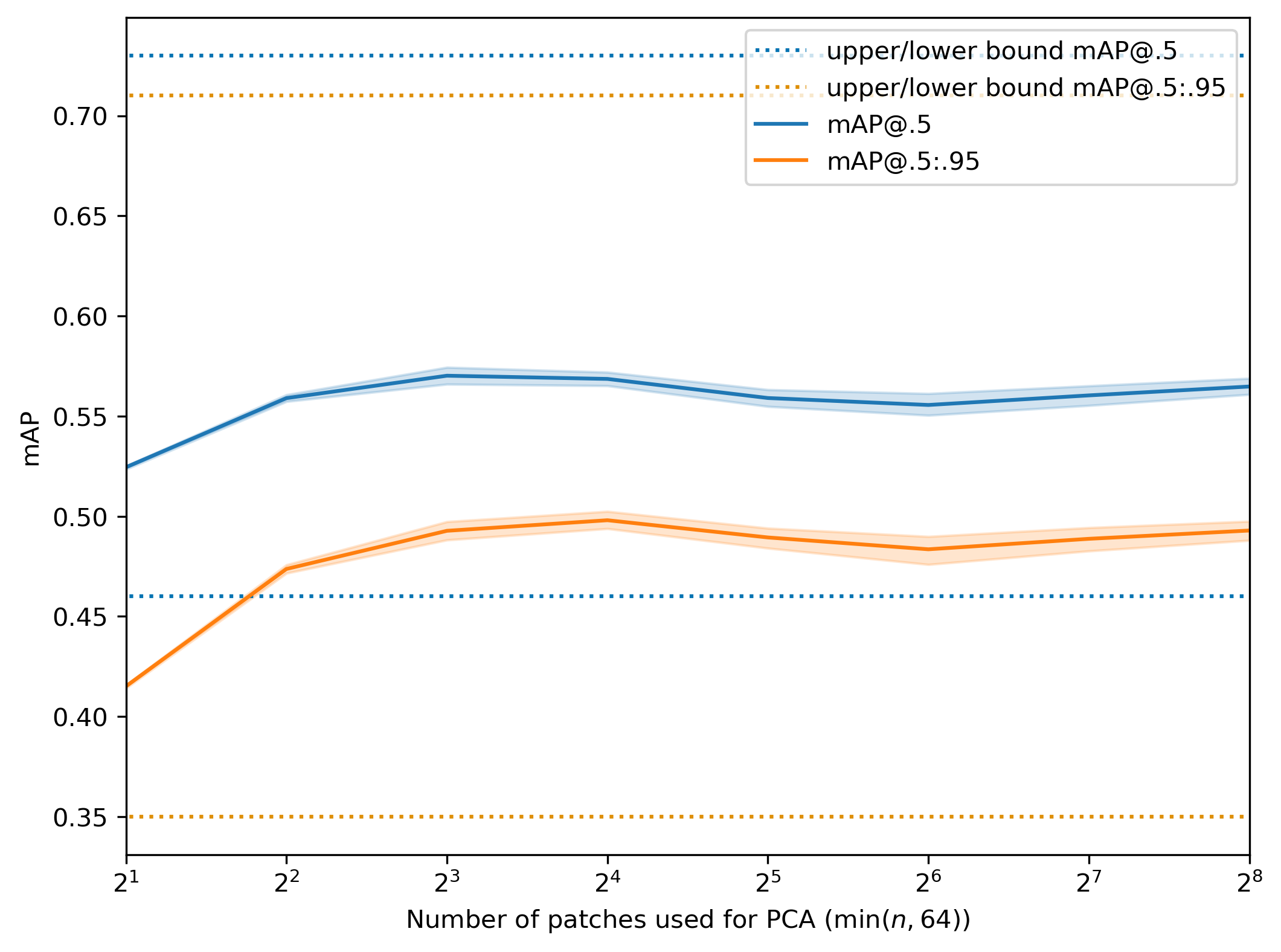}
    \caption{mAP@.5 and mAP@.5:.95 curve with different numbers of input elements for the PCA to recreate
    the patches with $\min(n,64)$ principal components.}
    \label{fig:mAPdrop_nelems}
\end{figure}

\begin{figure}
    \centering
    \includegraphics[width=0.8\textwidth]{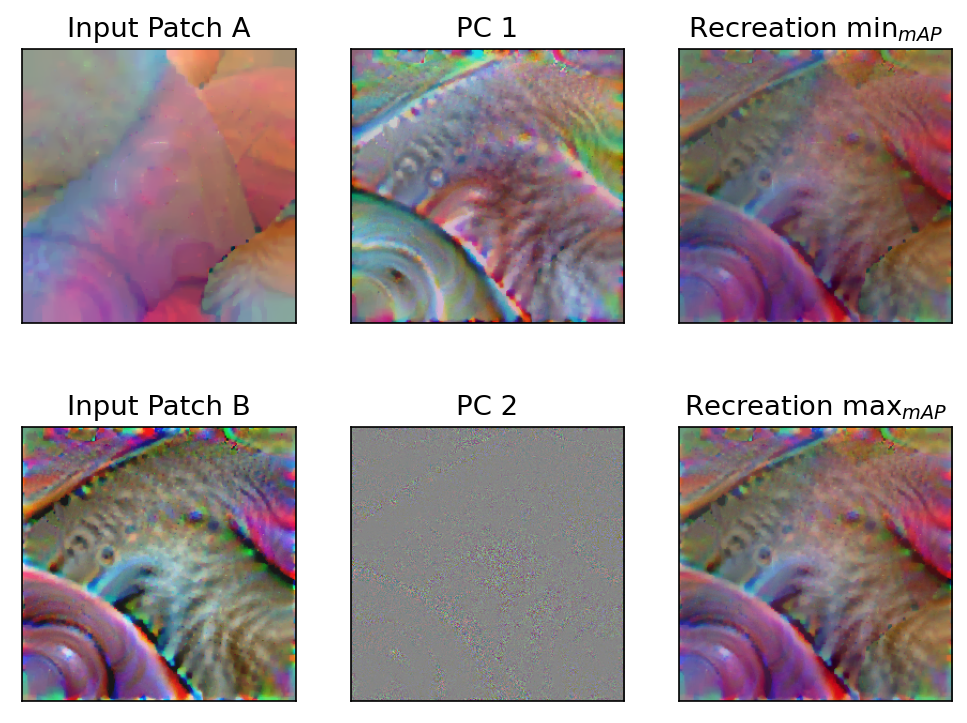}
    \caption{Left: The input set for the PCA. Middle: The computed principal components. Right: Example recreations of the patches by the two principal components.}
    \label{fig:PCA2Recreation}
\end{figure}

In comparison to \autoref{fig:mAPdrop}, the curves in \autoref{fig:mAPdrop_nelems} are not that intuitive. Here, the number 
of patches used to compute the PCA and the resulting mAPs are plotted. Since a PCA with $n$ components requires at least $n$
elements in the input set, the number of principal components is given by $min(n, 64)$. As a result, the number of principal
components used for this plot equals either the input set size or 64. 
As expected, the recreation of all 375 patches with only two components results in a small standard deviation of $0.01$ while
achieving a mean mAP:.5 of $0.52$, which differs from the PCA(256) by only a small percentage. A possible reason is, that with
such a small input set size, the resulting patch recreations do not vary a lot and reassemble, at best, one of the two
patches that are given in the input set. A look of the reassembled patches (see \autoref{fig:PCA2Recreation}) as well as the 
minimum achieved mAP values (0.41, 0.31) support this statement.


\section{Conclusion}
\label{sec:conclusion}
This paper investigates the applicability of the principal component analysis for adversarial patches 
to fool object detectors. The evaluation shows, that the recreation and sampling of patches based on 
the principal components is possible. As expected, those patches are generally not as good as carefully 
trained ones, yet they can be used as an initialization to finetune new patches. The evaluation also 
shows that the more principal components are used, the higher the mAP drop is. Nonetheless, the usage 
of the first eight principal components results already in a noticeable mAP drop.
The influence of the set size used for the PCA is after eight patches also quite stable.

Since these experiments are performed on a comparable small dataset and only a single object detector, future work should check the behavior 
with larger datasets (e.g., OpenImages \cite{Krasin2017}) and other object detectors (e.g. EfficientDet \cite{Tan2020} or DETR \cite{Carion2020}).
%
%
%
\section*{Acknowledgements}
This work was developed in Fraunhofer Cluster of Excellence \enquote{Cognitive Internet Technologies}.

 \bibliographystyle{splncs04}
 \bibliography{references}

\end{document}